\documentclass[runningheads]{llncs}
\usepackage[T1]{fontenc}
\usepackage{graphicx}
\usepackage{mwe}
\usepackage{float}
\usepackage{tikz,graphics,color,float,epsf,caption,subcaption}
\usepackage{enumitem}
\usepackage{subcaption}
\usepackage{amsmath}%
\usepackage{amsfonts}
\usepackage{multirow}%
\usepackage{xcolor}
\usepackage{booktabs}

\begin{document}
\title{Multi-Agent Reinforcement Learning for Autonomous Multi-Satellite Earth Observation: \\ A Realistic Case Study}
%
%
\author{Mohamad A. Hady\inst{1} \and Siyi Hu\inst{*,1} \and Mahardhika Pratama\inst{1} \and Jimmy Cao\inst{1} \and Ryszard Kowalczyk\inst{1,2}}

\institute{STEM, University of South Australia, Mawson Lakes, South Australia, Australia \email{mohamad.hady@mymail.unisa.edu.au}, \email{\{siyi.hu, dhika.pratama, jimmy.cao, ryszard.kowalczyk\}@unisa.edu.au} \and Systems Research Institute, Polish Academy of Sciences, Warsaw, Poland}

\authorrunning{Hady, et al.}
%
%

\maketitle              
\begin{abstract}
The exponential growth of Low Earth Orbit (LEO) satellites has revolutionised Earth Observation (EO) missions, addressing challenges in climate monitoring, disaster management, and more. However, autonomous coordination in multi-satellite systems remains a fundamental challenge. Traditional optimisation approaches struggle to handle the real-time decision-making demands of dynamic EO missions, necessitating the use of Reinforcement Learning (RL) and Multi-Agent Reinforcement Learning (MARL).
In this paper, we investigate RL-based autonomous EO mission planning by modelling single-satellite operations and extending to multi-satellite constellations using MARL frameworks. We address key challenges, including energy and data storage limitations, uncertainties in satellite observations, and the complexities of decentralised coordination under partial observability. By leveraging a near-realistic satellite simulation environment, we evaluate the training stability and performance of state-of-the-art MARL algorithms, including PPO, IPPO, MAPPO, and HAPPO. 
Our results demonstrate that MARL can effectively balance imaging and resource management while addressing non-stationarity and reward interdependency in multi-satellite coordination. The insights gained from this study provide a foundation for autonomous satellite operations, offering practical guidelines for improving policy learning in decentralised EO missions.

\end{abstract}
\section{Introduction}
The rapid expansion of Low Earth Orbit (LEO) satellites has significantly enhanced Earth Observation (EO) missions, driving advancements in climate monitoring, disaster response, agricultural planning, and urban development. However, managing multi-satellite constellations autonomously remains a fundamental challenge due to the dynamic and uncertain nature of space environments \cite{wang2020agile,chen2019mixedILP,stephenson2023optimal,pan2023dense}. Unlike traditional pre-planned missions, autonomous EO operations require satellites to make real-time decisions while managing resource constraints, operating under partial observability, and coordinating adaptively with other satellites \cite{li2024mission,Yang2024objective}. These challenges arise from several factors: uncertainties in observation conditions, such as variations in solar exposure affecting energy availability and imaging success; strict resource limitations on battery power, data storage, and reaction wheels, which demand efficient task scheduling; and non-stationarity in multi-agent environments, where each satellite's actions continuously alter system dynamics, complicating learning stability \cite{araguz2018applying,yao2019task}. Furthermore, the interdependence of satellite actions makes achieving mission objectives more complex, as operational success depends not only on individual decisions but also on the coordinated efforts of the entire constellation \cite{jun2021real,picard2021EOS}. Traditional optimisation techniques struggle to address these issues due to their reliance on predefined heuristics and limited adaptability to real-time operational demands, necessitating more advanced autonomous decision-making approaches.

Reinforcement Learning (RL) has emerged as a promising alternative for enabling autonomous decision-making in EO missions and has been applied to single-satellite scheduling, allowing satellites to balance imaging, energy consumption, and data management \cite{herrmann2024single,stephenson2024bsk}. However, as EO missions increasingly rely on multi-satellite constellations, the problem extends beyond single-agent decision-making to decentralised coordination among multiple satellites, which necessitates Multi-Agent Reinforcement Learning (MARL).
MARL provides a framework for decentralised decision-making in multi-satellite missions, where each satellite operates as an independent agent \cite{tang2024dynamic}. 
Early studies using centralised RL for satellite constellations \cite{herrmann2023reinforcement,stephenson2024reinforcement} struggle with scalability and rely on continuous communication, which is often impractical for real-world missions. To mitigate this, MARL-based frameworks have been proposed \cite{stephenson2024intent}, assuming ideal communication conditions where satellites continuously exchange information. However, in decentralised execution settings, such an approach imposes a heavy communication burden, as each satellite typically operates based only on its local observations.

\begin{figure}[t]
    \centering
    \includegraphics[width=\linewidth]{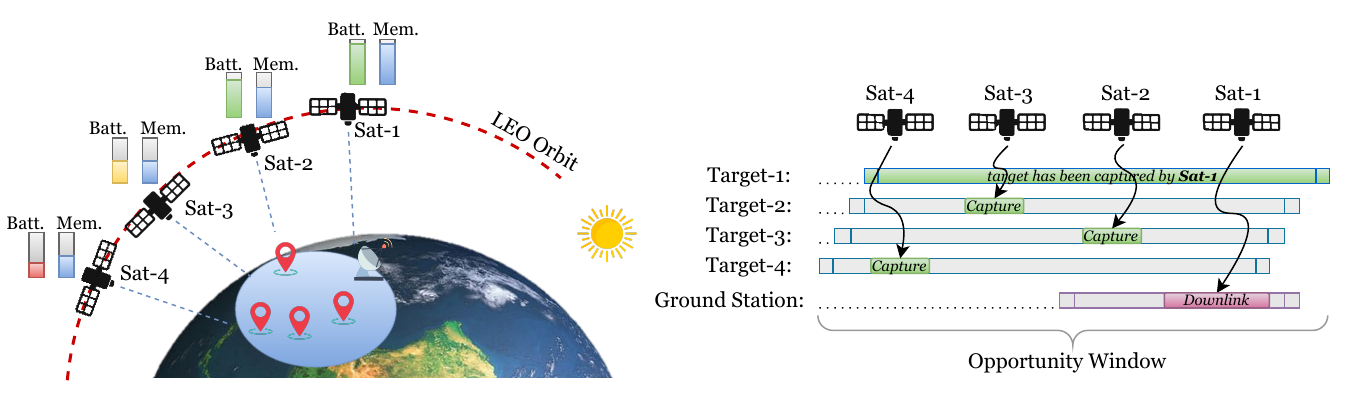}
    \caption{Multi-Satellite Cluster Image Capturing Task Scenario: Sat-1 to Sat-4 in a cluster constellation share the same four target opportunity windows. Sat-1, as the leading satellite, has the first access to the ground station and captures Target-1 in advance. The other satellites must capture different targets to ensure unique image captures. This behaviour introduces a non-stationarity issue in the multi-agent system. Each satellite has its own battery (Batt.) and data storage (Mem.) resources, which may be at different levels at the same time step $t$. This scenario highlights the importance of coordination and efficient resource management among satellites in autonomous EO missions.}
    \label{fig:problem}
\end{figure}

A variety of MARL learning frameworks have been developed to address these communication and coordination challenges, including Fully centralised (CTCE), Fully Decentralised (DTDE), and centralised Training Decentralised Execution (CTDE)\cite{ning2024survey}. CTCE relies on continuous information exchange, making it impractical for large-scale EO missions with limited communication. DTDE eliminates communication dependency but restricts coordination capabilities, as each satellite makes independent decisions. CTDE provides a balance by centralising training while enabling decentralised execution, allowing satellites to learn coordinated policies during training while operating autonomously based on local observations. Recent advancements, such as Multi-Agent Proximal Policy Optimisation (MAPPO) \cite{yu2022surprising} and Heterogeneous-Agent Proximal Policy Optimisation (HAPPO) \cite{kuba2022trust,zhong2024heterogeneous}, follow this paradigm, facilitating effective multi-satellite coordination under real-world constraints.

This work presents a comprehensive study of autonomous EO missions using RL and develops MARL within three different learning frameworks. 
Our contributions include: \textbf{1) A structured approach to modelling EO mission planning} extending the single-satellite with RL to multi-satellite coordination using MARL (see Fig.\ref{fig:problem}). This approach is inspired by realistic space system applications and incorporates satellite-specific operational factors such as varying state transition probabilities (e.g., eclipse conditions), action execution delays, uncertainty, and non-stationarity. \textbf{2) An in-depth investigation of learning stability and optimisation challenges} for both RL in single-satellite scenarios and MARL in multi-satellite EO missions. The study identifies critical factors, such as environmental dynamics and agent interactions, that contribute to unstable learning processes in these settings. \textbf{3) A comprehensive performance evaluation of state-of-the-art RL and MARL algorithms for autonomous EO missions}, comparing RL frameworks (e.g., PPO) for single-satellite tasks with MARL frameworks (e.g., Fully centralised PPO, Independent PPO (IPPO) \cite{de2020independent}, MAPPO, and HAPPO) for multi-satellite coordination. The analysis highlights the strengths and limitations of these methods in addressing scalability, coordination, and adaptability challenges. The code is made publicly available \footnotemark{}, with a demonstration video of our experimental scenario. \footnotetext{\url{https://anonymous.4open.science/r/Multi-Sat-MARL-2025}}

The rest of this paper is structured as follows: Section II presents the problem statement, starting from single-satellite to multi-satellite scenarios. Section III describes the algorithms employed to solve the problem. Section IV details our experimental evaluation and results. Finally, Section V concludes the paper and discusses future directions.

\section{Problem Statement}
In this section, a formal model of the Autonomous Multi-Satellite Earth Observation mission problem is discussed. The objective of the satellites is to capture as many unique images as possible during their orbits (see detail illustration in Fig \ref{fig:problem}. Previous works on single-satellite EO missions have formulated the problem as a sequential decision-making task in the well-known reinforcement learning framework, as a Partially Observable Markov Decision Process (POMDP) \cite{stephenson2024reinforcement,stephenson2024using}. Building upon this, we formally define the multi-satellite EO mission as a Decentralised POMDP (Dec-POMDP) model, extending from the single-agent MDP framework.

A single EO satellite functions as an agent, making decisions at discrete time steps based on its current state. The simulation defines four possible actions: \textit{1) Capturing the $i$-th image target}, where the satellite must orient its optical imaging sensor toward a selected target-$i$ on Earth and store it in the on-board memory; \textit{2) Downlinking}, where the satellite transmits the collected EO image data whenever it has access to a ground station; \textit{3) Charging}, which involves reorienting the satellite toward the sun to maximise solar energy absorption for recharging its battery; \textit{4) Desaturating}, which ensures that the Reaction Wheels (RWs), the primary actuators for attitude control, operate within safe operational limits.

To formally represent the autonomous satellite decision-making process as a POMDP, we define the problem as a tuple: \(\mathcal{G}=\langle\mathcal{S,A,O,T},r,\mathcal{Z,\gamma}\rangle,\)
where $\mathcal{S}$ represents a finite set of states that define the true underlying condition of the satellite in space, which is not fully observable. 
$\mathcal{A}$ is a finite set of available actions, including capturing, downlinking, charging, and desaturating. $\mathcal{O}$ is the finite set of observations the agent can receive, such as battery level, data storage availability, reaction wheel speed, target opportunity window, ground station access window, eclipse status, and time.
The transition probability function $\mathcal{T}(s_t,a_t,s_{t+1})=P(s_{t+1}|s_t,a_t)$ defines the probability of transitioning to state $s_{t+1}$ from $s_t$ after executing action $a_t$. The reward function $r$ determines the immediate reward for taking an action in a given state. The reward function for this EO mission is defined as:
\begin{equation}
    r(s_t,a_t)= 
    \begin{cases}
        \rho_j & \text{if unique image-$j$ is captured}, \\
        -100 & \text{if Failure occurs}, \\
        0 & \text{otherwise},
    \end{cases}
    \label{eq:reward}
\end{equation}
where $\rho_j \in [0,1]$ represents the reward based on the priority of target-$j$. This reward function is a crucial feedback to the agent indicating agent's policy performance. A negative reward is set 100 times greater than a successful target capturing to prevent agent entering failure condition. Ultimately, capturing many targets holds less value if it results a failure mode in the satellite. If the satellite encounters a failure, a fault condition is triggered, represented as:
\begin{equation}
    Failure = (b_t = 0 \space \vee \space \text{any}(\hat{\Omega}\geq\Omega_{max})).
    \label{eq:failure}
\end{equation}
The observation probability function $\mathcal{Z}(o_t,s_{t+1},a_t)=P(o_t|s_{t+1},a_t)$ defines the probability of observing $o_t$ given that the system is in state $s_{t+1}$ after taking action $a_t$. Finally, the discount factor $\mathcal{\gamma} \in [0,1]$ determines the importance of future rewards.

\subsection{Single-Satellite Problem}
\textbf{Constraints and Limitations:} The satellite has two limited resources that are considered \textit{constraints} in our study: battery level ($b_t \in [B_{min},B_{max}]$) and data storage capacity ($d_t \in [D_{min},D_{max}]$) at any time step ($t$). At each time step, the satellite consumes electrical power, denoted as $c_{b,i}$, and stores data, represented as $c_{d,i}$. To maximise battery charging, the satellite must adjust its attitude toward the sun, which may conflict with its target imaging orientation.
Another constraint arises from attitude control, specifically the speed of the Reaction Wheels (RWs), denoted as $\hat{\Omega} \in [-\Omega_{max},\Omega_{max}]$. These wheels serve as the primary actuators for satellite attitude adjustments along the three axes ($x,y,z$). To prevent exceeding the maximum speed threshold, the satellite must periodically \textit{Desaturate} the wheels. The limited resource constraints are mathematically expressed as: \(\sum^{\infty}_{t=0}{c_{b,t}}\leq b_t, \: B_{min}\leq b_t \leq B_{max} \quad\) and \(\quad \sum^{\infty}_{t=0}{c_{d,t}}\leq d_t, \: D_{min}\leq d_t \leq D_{max}.\) 
These constraints are incorporated into the model as \textit{Failure} (Eq. \ref{eq:failure}) in the reward function, resulting in penalties or negative rewards. Some other constraint such as communication baud rate has relevant impact to the system performance, however it is not a major element causing satellite fault condition.

\textbf{Uncertainties and Randomness:} 
Real-world systems inherently involve uncertainties due to noise, disturbances, and unpredictable variations. In EO satellite missions, such uncertainties are closely tied to resource availability, which influences both the observation probability function and action necessity.
Initial battery levels affect learning stability, as charging is only possible in unshaded regions. Likewise, limited initial data storage requires the policy to balance imaging and downlinking to maintain memory for future captures. Thus, initial resource availability directly impacts observation quality and action feasibility, playing a key role in the POMDP formulation. The POMDP tuple definition ($\mathcal{G}$) inherently captures these uncertainties, as the satellite lacks full state observability during operation.

\subsection{Multi-Satellite Problem}
\textbf{Multi-Satellite Coordination:} In \textit{cooperative} MARL, agents share a global objective and optimise a common reward function (Eq. \ref{eq:reward}). This work adopts the \textit{Decentralised Partially Observable Markov Decision Process (Dec-POMDP)} framework, well-suited for multi-satellite EO missions where each agent acts based on local observations \cite{oliehoek2016concise}.
A Dec-POMDP is defined by the tuple: \(\mathcal{D}=\langle S, \{A_i\}_{i=1}^{N}, T, r, \{O_i\}_{i=1}^{N}, O, N, \gamma \rangle,\)
where $S$ represents the set of environment states, and each agent $i$ has an action space $A_i$, forming the joint action space $A = A_1 \times \dots \times A_N$ for $N$ agents. The state transition function $T: S \times A \times S \rightarrow [0,1]$ describes the probability of transitioning from state $s$ to $s'$ given the joint action $\mathbf{a} = (a_1, \dots, a_N)$. The global reward function $r: S \times A \rightarrow \mathbb{R}$ provides feedback based on joint actions. Each agent $i$ has an observation space $O_i$, and the joint observation space is $O = O_1 \times \dots \times O_N$. The observation function $O: S \times A \times O \rightarrow [0,1]$ defines the probability of an agent receiving observation $o_i$ given state $s$ and joint action $\mathbf{a}$. The number of agents is $N$, and $\gamma \in [0,1]$ is the discount factor, which controls the importance of future rewards. Since Dec-POMDPs operate under decentralised execution, each satellite makes decisions based only on its partial observations while still contributing to the shared objective.
This decentralised yet cooperative nature introduces challenges in coordinating image captures among satellites, as they must infer the actions of others without direct communication. 

\textbf{Non-Stationarity:} Non-stationarity in multi-satellite settings arises because the environment is constantly evolving as agents update their policies during training. 
As one agent updates its policy, it alters the environment for others, causing the distribution of states and actions to shift. This violates the stationarity assumption in most reinforcement learning algorithms, leading to instability in training. Another challenge is the \textit{reward interdependency}, where a satellite's reward is influenced by the actions of others. For example, the global reward based on unique image captures depends on the joint action vector $\mathbf{a} = (a_1, a_2, \dots, a_N)$. As satellites adjust their policies, the reward landscape changes, increasing the complexity of policy learning.

\section{Methods}
To solve POMDP and Dec-POMDP of Autonomous EO Mission, a model-free approach is selected due to its flexibility to directly learn the policy, especially when the model of the system is complex and highly dynamic. Hence, RL is initially tailored to handle single-satellite EO mission then followed by the extension to MARL with multi-satellite constellation.

\subsection{Reinforcement Learning}
A single-satellite has an on-board processor to execute the best autonomy policy. Here, the satellite is defined as an agent to solve the satellite decision making during performing EO Mission. RL is used to learn an optimal policy ($\pi^*$) which maps states to actions $\pi(a_t|s_t)$ and maximises the expected cumulative reward, or return, over time \cite{sutton2018reinforcement}. The return at time step \( t \), denoted \( R_t \), is expressed as the sum of discounted rewards \( R_t = \sum_{i=0}^{\infty} \gamma^i r_{t+i+1}\). To measure how good an observation is under a particular policy, a State-Value function is defined as the expected cumulative reward obtained from Eq. \ref{eq:reward}: \(V_{\pi}(s)=\mathbb{E}_{\pi}\left[ R_t |S_t=s\right]\) and an Action-Value function: \(Q^\pi(s, a) = \mathbb{E}_\pi \left[ R_t | S_t = s, A_t = a \right]\).

\textbf{Proximal Policy Optimisation:} Among the variety of the RL approaches, our work focus on the \textit{on-policy methods} which is Proximal Policy Optimisation (PPO) \cite{schulman2017PPO}. This algorithm is a widely used reinforcement learning (RL) algorithm, particularly for solving high-dimensionality problems. It belongs to the family of policy optimisation methods with actor-critic networks architecture, where the goal is to optimise the policy directly instead of learning a value function. PPO aims to improve the policy $\pi_{\theta}(a|s)$, parameterised by $\theta$, to maximise the expected cumulative reward: $J(\theta) = \mathbb{E}_{\tau \sim \pi_\theta} \left[ \sum_{t=0}^\infty \gamma^t r_t \right]$. Instead of directly optimising $J(\theta)$, PPO uses a surrogate objective function to limit policy updates and stabilise the training phase:

\begin{equation}
    L^{\text{PPO}}(\theta) = \mathbb{E}_t \left[ \min\left(r_t(\theta) \hat{A}_t, \text{clip}(r_t(\theta), 1 - \epsilon, 1 + \epsilon) \hat{A}_t\right) \right],
\end{equation}
where $r_t(\theta)=\frac{\pi_{\theta}(a|s)}{\pi_{\theta_{old}}(a|s)}$ is the probability ratio between the new and old policies. $\hat{A}_t$ is the estimated advantage function, typically calculated using the Generalised Advantage Estimation (GAE). And $\epsilon$ is a hyperparameter controlling the clipping range. The "clip" term ensures that the probability ratio $r_t(\theta)$ does not deviate too much from 1, avoiding overly large policy updates. This stabilises training and prevents performance collapse, which is more suitable for a realistic satellite simulation with uncertainty and randomness. 
PPO also includes a value function loss to improve the policy's state value predictions: $L^{\text{value}}(\theta) = \mathbb{E}_t \left[ \left( V_\theta(s_t) - V_t^\text{target} \right)^2 \right]$. Then, to encourage exploration, an entropy bonus term is added: $L^{\text{entropy}}(\theta) = \mathbb{E}_t \left[ -\pi_\theta(a_t | s_t) \log \pi_\theta(a_t | s_t) \right]$. The total loss function for PPO is a combination of the surrogate objective, value function loss, and entropy bonus: $L^{\text{total}}(\theta) = L^{\text{PPO}}(\theta) - c_1 L^{\text{value}}(\theta) + c_2 L^{\text{entropy}}(\theta)$, where $c_1$ and $c_2$ are the coefficients used for balancing the contributions of the different terms.

\subsection{Multi-Agent Reinforcement Learning}
Effective multi-satellite constellation management requires cooperative control to optimise global performance and enhance EO missions. MARL extends RL to multi-agent settings, allowing agents to learn and interact simultaneously \cite{bu2008comprehensive}. Autonomous decision-making can be executed in a single resource (a processing satellite or ground station) or be distributed via onboard processing. 
\textbf{Centralised Training Decentralised Execution (CTDE) Learning Paradigm:} 
This framework strikes a balance between leveraging centralised information during training and enabling decentralised decision-making during execution. Moreover, it enables scalability feature and real-world applicability. Each agent $i \in \mathcal{N}$ learns a policy $\pi_i(a_i|o_i)$, where $o_i$ is the local observation of agent-$i$. The training uses a centralised critic $Q_{\theta}(o,\boldsymbol{a})$ that evaluates joint actions based on the global observation $\boldsymbol{o}$: $Q_\theta(s, \mathbf{a}) = \mathbb{E} \left[ \sum_{t=0}^\infty \gamma^t r(s_t, \mathbf{a}_t) \mid s_0 = s, \mathbf{a}_0 = \mathbf{a} \right]$. And during execution, the policy of each agent is still in decentralised manner: $\pi_i(a_i|o_i)$. In our works, two recent state of the art in CTDE has been selected to be compared with fully centralised and decentralised algorithm: 

\textit{1) Multi-agent PPO (MAPPO)} is an extension of PPO designed specifically for multi-agent systems \cite{yu2022surprising}. It incorporates centralised critics and decentralised policies to improve performance in MARL tasks. MAPPO uses a single centralised critic shared by all agents, allowing the evaluation of the global state to stabilise learning and mitigate non-stationarity: $V_i^{\text{centralised}}(s) \approx \mathbb{E} \left[ \sum_{t=0}^\infty \gamma^t r_{i,t} \mid s_0 = s \right]$, where \( s \) represents the global state, \( \gamma \) is the discount factor, and \( r_t \) is the reward at time step \( t \). The loss function for the policy optimisation in MAPPO is given by: 
\begin{equation}
    L^{\text{MAPPO}}(\theta_i) = \mathbb{E}_t \left[ \min\left(r_t(\theta_i) \hat{A}_t, \text{clip}(r_t(\theta_i), 1-\epsilon, 1+\epsilon) \hat{A}_t\right) \right],
\end{equation}
where: \( r_t(\theta_i) = \frac{\pi_{\theta_i}(a_t \mid o_t)}{\pi_{\theta_i}^{\text{old}}(a_t \mid o_t)} \) is the probability ratio,\( \hat{A}_t \) is the global advantage function, and \( \epsilon \) is the clipping parameter.

\textit{2) Heterogeneous Agent PPO (HAPPO)} extends MAPPO by accounting for heterogeneous agents with distinct state-action spaces or roles and the sequential update scheme \cite{zhong2024heterogeneous}. It uses individual advantage functions and decentralised policies while maintaining centralised critics.
In HAPPO, the centralised value function is agent-specific to handle heterogeneous agents: $V_i^{\text{centralised}}(s) \approx \mathbb{E} \left[ \sum_{t=0}^\infty \gamma^t r_{i,t} \mid s_0 = s \right]$, where \( i \) denotes the agent index and \( r_{i,t} \) is the reward specific to agent \( i \). And, the loss function for HAPPO is denoted by:
\begin{equation}
    L_i^{\text{HAPPO}}(\theta_i) = \mathbb{E}_t \left[ \min\left(r_{i,t}(\theta_i) \hat{A}_{i,t}, \text{clip}(r_{i,t}(\theta_i), 1-\epsilon, 1+\epsilon) \hat{A}_{i,t}\right) \right],
\end{equation}
where \( r_{i,t}(\theta_i) = \frac{\pi_{\theta_i}(a_{i,t} \mid o_{i,t})}{\pi_{\theta_i}^{\text{old}}(a_{i,t} \mid o_{i,t})} \) and \( \hat{A}_{i,t} \) is the advantage function of agent \( i \).

Both MAPPO and HAPPO operate effectively in the CTDE paradigm. It has main differentiations in the centralised critics, where MAPPO uses a single global critic for shared evaluation and HAPPO incorporates agent-specific critics for greater flexibility in heterogeneous systems. It shares the same decentralised execution settings, where each agent executes its policy based solely on local observations, making these algorithms suitable for real-world scenarios where the global state information is unavailable during execution.

\section{Experimental Results}
\label{section4:experiment}
Our experiments evaluate EO mission performance to capture as many unique images as possible from 2,000 uniformly distributed targets around the world within 2 times orbit. Our experiments consist of two parts: (1) Single-agent RL is applied to a single-satellite to assess RL effectiveness, and (2) MARL is introduced by scaling up to a four-satellite constellation. Adjustable satellite parameters in BSK-RL are listed in Supplemental Document. In multi-satellite experiment, we evaluate the Walker-delta and Cluster orbit types. 

\subsection{Single-Satellite RL}
\label{Section411:limited}
In our experiments, to demonstrate the \textbf{limited resources} availability problem, the battery and data storage capacity has been defined as: $B = (50,400)$ Wh, of the Battery and $D=(5,500)$ GB of the Data Storage for a single satellite. The transmitter baud rate it is defined as $Bdr=(0.5,4.3)$ Mbps and captured image size $img=(Small(S),Large (L))$. The result is provided in Supplemental Document Fig 1. left side, generally, the higher amount of resources capacity has less challenge to the PPO learning performance. The limited data storage resources reduces the learning performance more than others. A limited battery resource causes multiple significant drops in the learning performance, since it triggers the failure penalty.

Also, we evaluate PPO performance under various sources of \textbf{Uncertainty and Randomness Challenge} (see Supplemental Document Fig. 1). Randomness introduces learning fluctuations, with the following settings: attitude disturbance (normally distributed, scale $10^{-4}$ for yaw, pitch, roll), initial reaction wheel (RW) speeds uniformly in -3000,3000 Rpm, battery level in 40-80\%, and data storage in 20-80\%.
Random RW speeds notably degrade maneuver accuracy. Variations in initial data storage pose decision-making challenges and destabilise learning. In contrast, battery level randomness has a minor effect, and attitude disturbances only affect image capture without penalty. The greatest challenge arises when RW speed and data storage are both highly randomised, significantly hindering policy convergence.

\subsection{Multi-Satellite MARL} 
Our experiment focuses on cooperative satellites observing shared targets and compensating for each other's constraints, such as limited energy or storage. Then, the performance of MARL algorithms, including the Centralised PPO, IPPO (decentralised), MAPPO and HAPPO, under limited resources challenges, uncertainty and randomness. The identification of effective approaches for enhancing autonomous collaboration decision-making in dynamic, resource-constrained satellite missions is examined by comparing algorithm performance.  

\begin{figure}[t]
    \centering
    \includegraphics[width=\linewidth]{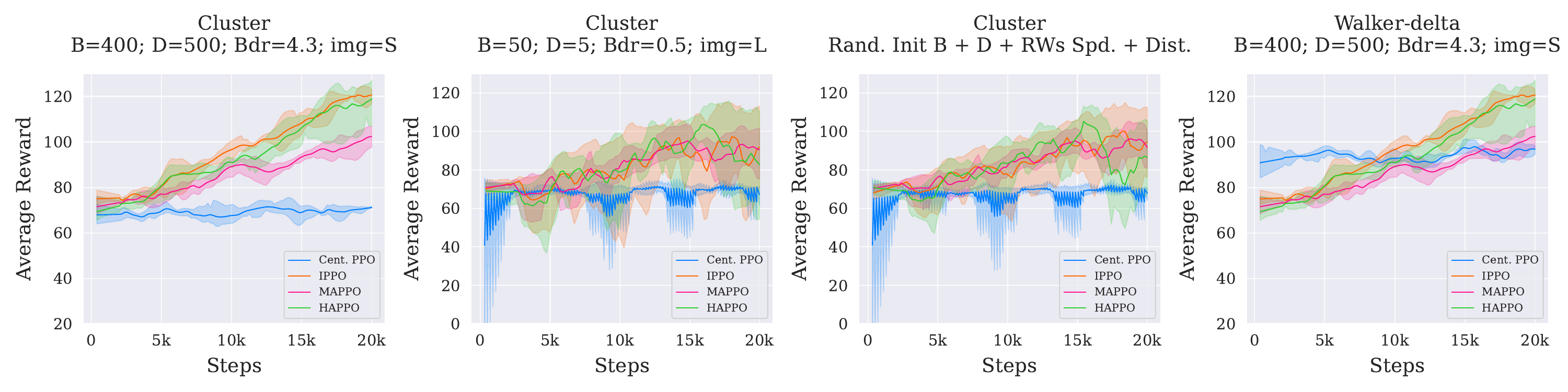}
    \caption{Multi-Satellite Learning Performance Under Cluster and Walker-Delta Orbits: Evaluated with both default and limited resources, including Battery ($B$), Data Storage ($D$), Baud Rate ($Bdr$), captured image sizes ($img$), and the presence of randomness.}
    \label{fig:multiSatAll}
\end{figure}

\begin{figure}[t]
    \centering
    \begin{subfigure}[b]{0.23\textwidth}
        \centering
        \includegraphics[width=\linewidth]{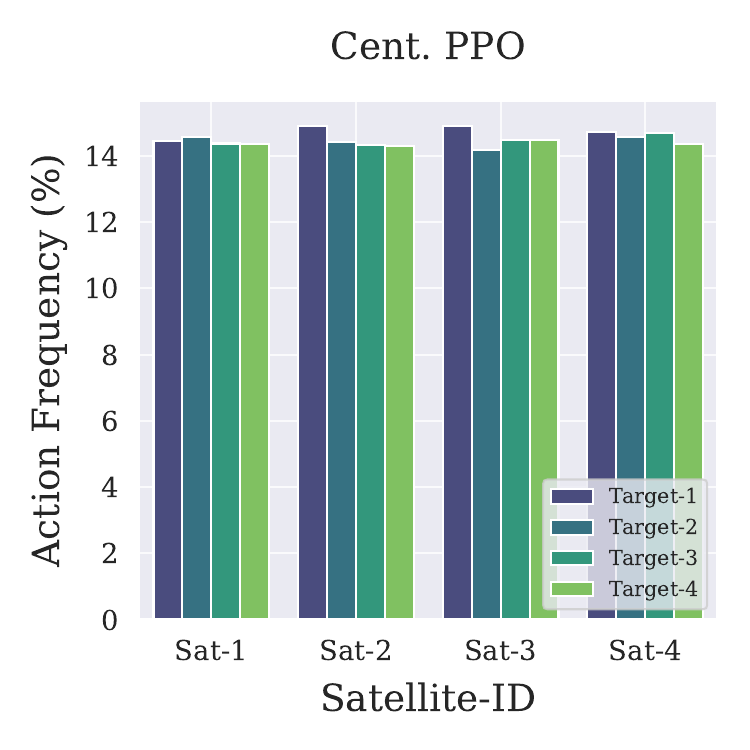}
    \end{subfigure}
    \hfill
    \begin{subfigure}[b]{0.23\textwidth}
        \centering
        \includegraphics[width=\linewidth]{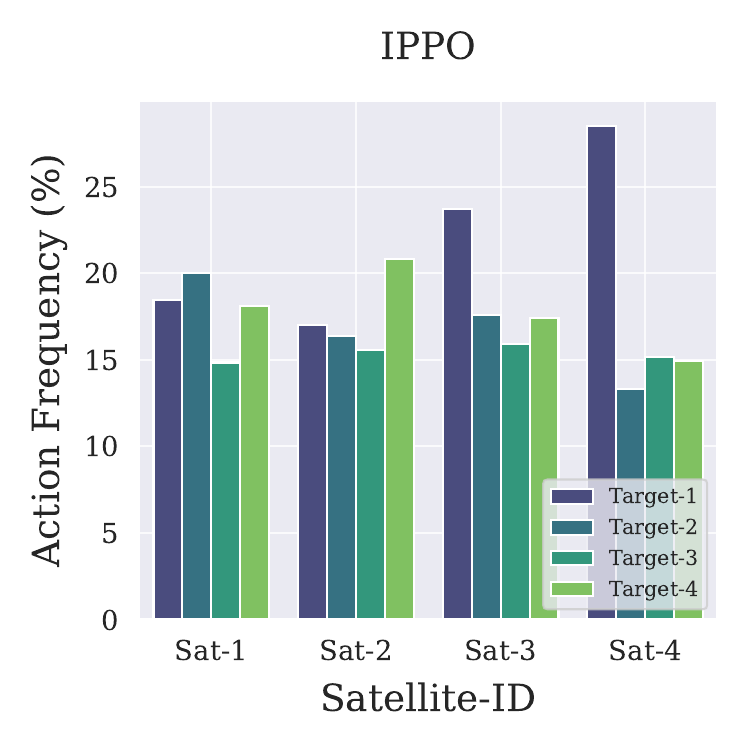}
    \end{subfigure}
    \hfill
    \begin{subfigure}[b]{0.23\textwidth}
        \centering
        \includegraphics[width=\linewidth]{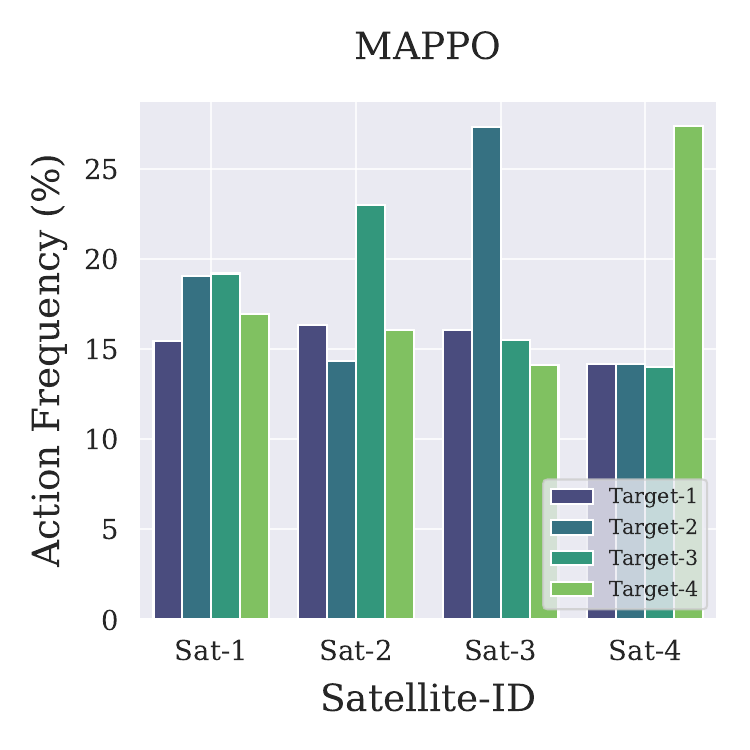}
    \end{subfigure}
    \hfill
    \begin{subfigure}[b]{0.23\textwidth}
        \centering
        \includegraphics[width=\linewidth]{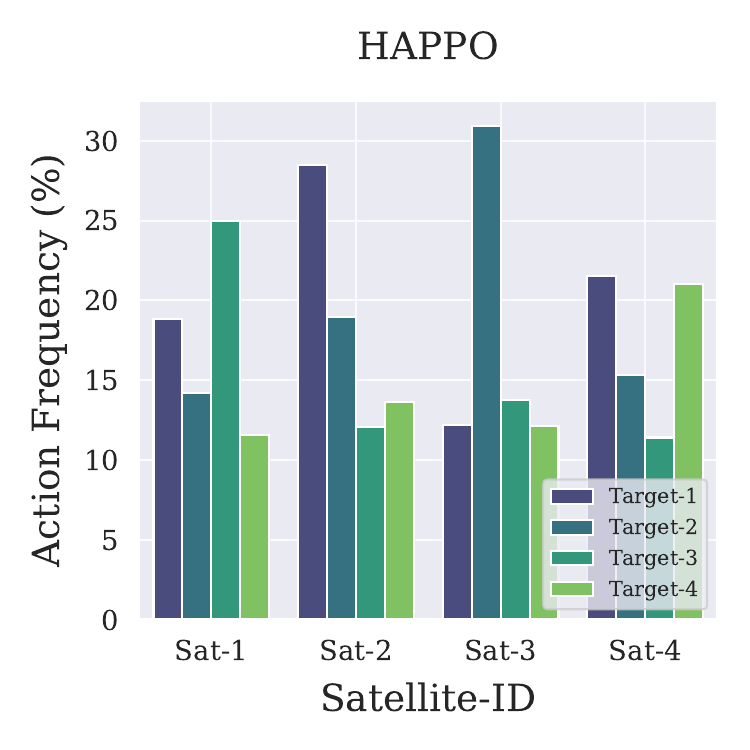}
    \end{subfigure}
    \caption{Target Capturing Action Frequencies Across Different Satellites and Algorithms: Evaluated under varying data storage capacities ($D$), with Sat-1 to Sat-4 having (5, 10, 250, 500) GB.}
    \label{fig:MARLcoordination_mem}
\end{figure}

\textbf{Limited Resources Capacity:} The resources specification used in this study has the same definition as in Section \ref{Section411:limited}. Learning performance of multi-satellite under limited resources are presented in Fig. \ref{fig:multiSatAll}, the second column from the left side. The Centralised PPO algorithm performance falls into a sub-optimal condition easily since it suffers from the non-stationarity problem. Besides, IPPO has competitive results yet slightly under the MAPPO and HAPPO performance. And, HAPPO has a better performance while it is implemented under large resources capacity. The behaviour of the presence of limited resources challenges multi-satellite is pretty similar to the single-satellite.

\textbf{Uncertainty and random initialisation:} Experimental results under uncertainty and randomness (see the third left column in  Fig. \ref{fig:multiSatAll}) revealed that centralised methods performed poorly due to their sensitivity to environmental variability, such as reaction wheel speed initialisation and data storage randomness. The other algorithms, IPPO, MAPPO, and HAPPO, demonstrate similar behaviour and more stable performance, effectively adapting to the fluctuating conditions. These algorithms show resilience in handling uncertainties, maintaining consistent performance despite the challenges, highlighting the advantages of decentralised training in dynamic, resource-constrained environments.

\textbf{Walker-delta and Cluster Constellation:} In different constellation settings, Walker-delta is considered as less-complex due to it has low collaboration properties yielding a weak non-sationarity issue compared to cluster. 
At the beginning, training phase in Walker-delta (Fig. \ref{fig:multiSatAll} most right side column), the Centralised PPO has a higher rewards and keep increasing slightly and still under the other algorithms performance at 20k steps. However, the IPPO, HAPPO shares a similar learning performance in this settings. 

\textbf{Multi-Satellite Coordination in Diverse Satellite Resources:} In multi-satellite coordination, capturing actions across satellites are analysed to assess collaboration capability of MARL algorithms (see Fig. \ref{fig:MARLcoordination_mem}). To encourage cooperation, we assign varying data storage capacities: Sat-1 to Sat-4 have 5, 10, 250, and 500 GB, respectively. Satellites with limited storage tend to capture fewer targets, while those with more resources capacity compensate for them. Our results demonstrate the policy has coordination behaviour to capture different targets, highlighting the MARL’s effectiveness. The Centralised PPO relies on single policy strategy and it has similar policy for any satellites leading to similar action characteristics and low coordination. The IPPO learns multiple policies independently and has more capture for Sat-3 and Sat-4, but they cover the same Target-1. MAPPO policy has the best coordination properties, different satellite focus on different Targets. HAPPO has coordination properties as different satellites has different target preferences and covers less resource capacity satellite. However, Sat-4 has duplication on Target-1 and Target-4, where both share a similar action frequency.

\section{Conclusion}
This study implemented the RL and its extension to MARL algorithms in the realistic environment of autonomous EO mission. Through extensive experiments under various uncertainties, such as: reaction wheel initialisation, data storage, and attitude disturbances, it has demonstrated the effectiveness and adaptability of CTDE-MARL, in addressing the challenges of dynamic and resource-constrained Earth observation missions. Moreover, the results highlight the potential of MAPPO in MARL frameworks to enhance collaboration and retain the communication efficiency in satellite operations. In the future, this work can be extended further to explore more complex scenarios, such as heterogeneous satellite constellations with diverse sensor capabilities and larger-scale of multi-cluster orbit systems. Additionally, integrating domain-specific knowledge into MARL training and developing methods to further mitigate non-stationarity issue will be key directions to improve real-world deployment feasibility.

\begin{credits}
\subsubsection{\ackname} This work has been supported by the SmartSat CRC, whose activities are funded by the Australian Government’s CRC Program. This work use an open-source realistic satellite simulator (Basilisk and BSK-RL) that is actively developed by Dr. Hanspeter Schaub and team at AVS Laboratory, University of Colorado Boulder. Also, the authors would like to express their sincere gratitude to BAE Systems for their invaluable support and collaboration throughout this research.
\end{credits}

\bibliographystyle{ieeetr}
\bibliography{references}

@article{wang2020agile,
  title={Agile earth observation satellite scheduling over 20 years: Formulations, methods, and future directions},
  author={Wang, Xinwei and Wu, Guohua and Xing, Lining and Pedrycz, Witold},
  journal={IEEE Systems Journal},
  volume={15},
  number={3},
  pages={3881--3892},
  year={2020},
  publisher={IEEE}
}

@article{jun2021real,
  title={Real-time online rescheduling for multiple agile satellites with emergent tasks},
  author={Jun, Wen and Xiaolu, Liu and Lei, He},
  journal={JSEE},
  volume={32},
  number={6},
  pages={1407--1420},
  year={2021},
  publisher={BIAI}
}

@inproceedings{stephenson2023optimal,
  title={Optimal Target Sequencing in the Agile Earth-Observing Satellite Scheduling Problem Using Learned Dynamics},
  author={Stephenson, Mark and Schaub, Hanspeter},
  booktitle={The AAS/AIAA Astrodynamics Specialist Conference, Big Sky, MT, USA},
  pages={13--17},
  year={2023}
}

@article{li2024mission,
  title={Mission planning for distributed multiple agile Earth observing satellites by attention-based deep reinforcement learning method},
  author={Li, Peiyan and Wang, Huiquan and Zhang, Yongxing and Pan, Ruixue},
  journal={Advances in Space Research},
  year={2024},
  publisher={Elsevier}
}

@article{chen2019mixedILP,
  title={A mixed integer linear programming model for multi-satellite scheduling},
  author={Chen, Xiaoyu and Reinelt, Gerhard and Dai, Guangming and Spitz, Andreas},
  journal={European Journal of Operational Research},
  volume={275},
  number={2},
  pages={694--707},
  year={2019},
  publisher={Elsevier}
}

@article{ning2024survey,
  title={A survey on multi-agent reinforcement learning and its application},
  author={Ning, Zepeng and Xie, Lihua},
  journal={Journal of Automation and Intelligence},
  year={2024},
  publisher={Elsevier}
}

@inproceedings{stephenson2024bsk,
  title={BSK-RL: Modular, High-Fidelity Reinforcement Learning Environments for Spacecraft Tasking},
  author={Stephenson, Mark A and Schaub, Hanspeter},
  booktitle={75th International Astronautical Congress, Milan, Italy, IAF},
  year={2024}
}

@inproceedings{stephenson2024using,
  title={Using Enhanced Simulation Environments to Accelerate Reinforcement Learning for Long-Duration Satellite Autonomy},
  author={Stephenson, Mark and Mantovani, Lorenzzo and Phillips, Sean and Schaub, Hanspeter},
  booktitle={AIAA SCITECH 2024 Forum},
  pages={0990},
  year={2024}
}

@inproceedings{stephenson2024reinforcement,
  title={Reinforcement Learning For Earth-Observing Satellite Autonomy With Event-Based Task Intervals},
  author={Stephenson, Mark and Schaub, Hanspeter},
  booktitle={AAS Rocky Mountain GN\&C Conference, Breckenridge, CO},
  year={2024}
}

@article{herrmann2024single,
  title={Single-Agent Reinforcement Learning for Scalable Earth-Observing Satellite Constellation Operations},
  author={Herrmann, Adam and Stephenson, Mark A and Schaub, Hanspeter},
  journal={Journal of Spacecraft and Rockets},
  volume={61},
  number={1},
  pages={114--132},
  year={2024},
  publisher={American Institute of Aeronautics and Astronautics}
}

@inproceedings{herrmann2023reinforcement ,
  title={Reinforcement learning for multi-satellite agile earth observing scheduling under various communication assumptions},
  author={Herrmann, A and Stephenson, M and Schaub, H},
  booktitle={AAS Rocky Mountain GN\&C Conference},
  year={2023}
}

@inproceedings{stephenson2024intent,
  title={Intent Sharing for Emergent Collaboration in Autonomous Earth Observing Constellations},
  author={Stephenson, Mark and Mantovani, Lorenzzo and Schaub, Hanspeter},
  booktitle={AAS Guidance and Control Conference, Breckenridge, CO},
  pages={24--192},
  year={2024}
}

@book{oliehoek2016concise,
  title={A concise introduction to decentralized POMDPs},
  author={Oliehoek, Frans A and Amato, Christopher and others},
  volume={1},
  year={2016},
  publisher={Springer}
}

@book{sutton2018reinforcement,
  title={Reinforcement learning: An introduction},
  author={Sutton, Richard and Barto, Andrew},
  year={2018},
  publisher={MIT press}
}

@article{schulman2017PPO,
  title={Proximal policy optimization algorithms},
  author={Schulman, et.,al.},
  journal={arXiv preprint arXiv:1707.06347},
  year={2017}
}

@article{bu2008comprehensive,
  title={A comprehensive survey of multiagent reinforcement learning},
  author={Bu, Lucian and Babu, Robert and De Schutter, Bart and others},
  journal={IEEE-TSMC. C, Appl., Rev.},
  volume={38},
  number={2},
  pages={156--172},
  year={2008},
  publisher={IEEE}
}

@article{yu2022surprising,
  title={The surprising effectiveness of ppo in cooperative multi-agent games},
  author={Yu, Chao and Velu, Akash and Vinitsky, Eugene and Gao, Jiaxuan and Wang, Yu and Bayen, Alexandre and Wu, Yi},
  journal={Advances in Neur-IPS},
  volume={35},
  pages={24611--24624},
  year={2022}
}

@inproceedings{kuba2022trust,
    title={Trust Region Policy Optimisation in Multi-Agent Reinforcement Learning},
    author={Jakub Grudzien Kuba and Ruiqing Chen and Muning Wen and Ying Wen and Fanglei Sun and Jun Wang and Yaodong Yang},
    booktitle={ICLR},
    year={2022}
}

@article{zhong2024heterogeneous,
  title={Heterogeneous-agent reinforcement learning},
  author={Zhong, Yifan and Kuba, Jakub Grudzien and Feng, Xidong and Hu, Siyi and Ji, Jiaming and Yang, Yaodong},
  journal={JMLR},
  volume={25},
  pages={1--67},
  year={2024}
}

@article{de2020independent,
  title={Is independent learning all you need in the starcraft multi-agent challenge?},
  author={De Witt, Christian Schroeder and Gupta, Tarun and Makoviichuk, Denys and Makoviychuk, Viktor and Torr, Philip HS and Sun, Mingfei and Whiteson, Shimon},
  journal={arXiv preprint arXiv:2011.09533},
  year={2020}
}

@article{araguz2018applying,
  title={Applying autonomy to distributed satellite systems: Trends, challenges, and future prospects},
  author={Araguz, Carles and Bou-Balust, Elisenda and Alarc{\'o}n, Eduard},
  journal={Systems Engineering},
  volume={21},
  number={5},
  pages={401--416},
  year={2018},
  publisher={Wiley Online Library}
}

@article{yao2019task,
  title={Task allocation strategies for cooperative task planning of multi-autonomous satellite constellation},
  author={Yao, Feng and Li, Jiting and Chen, Yuning and Chu, Xiaogeng and Zhao, Bang},
  journal={Advances in space research},
  volume={63},
  number={2},
  pages={1073--1084},
  year={2019},
  publisher={Elsevier}
}

@article{tang2024dynamic,
  title={A Dynamic and Collaborative Spectrum Sharing Strategy Based on Multi-Agent DRL in Satellite-Terrestrial Converged Networks},
  author={Tang, Chao and Chen, Yueyun and Chen, Guang and Du, Liping and Liu, Huan},
  journal={IEEE Transactions on Vehicular Technology},
  year={2024},
  publisher={IEEE}
}

@inproceedings{picard2021EOS,
author = {Picard, Gauthier and Caron, Cl\'{e}ment and Farges, Jean-Loup and Guerra, Jonathan and Pralet, C\'{e}dric and Roussel, St\'{e}phanie},
title = {Autonomous Agents and Multiagent Systems Challenges in Earth Observation Satellite Constellations},
year = {2021},
isbn = {9781450383073},
publisher = {IFAAMAS},
address = {Richland, SC},
booktitle = {Proceedings of the 20th AAMAS Conference},
pages = {39–44},
numpages = {6},
location = {Virtual Event, United Kingdom},
series = {AAMAS '21}
}

@inproceedings{Yang2024objective,
author = {Yang, Xueying and Hu, Min and Huang, Gang},
title = {Objective task matching strategy for Multi-Satellite Imaging Mission Planning in complex heterogeneous scenarios},
year = {2024},
isbn = {9798400709258},
publisher = {ACM},
address = {New York, NY, USA},
doi = {10.1145/3638264.3638284},
pages = {96–101},
numpages = {6},
location = {Chengdu, China},
series = {MICML '23}
}

@inproceedings{pan2023dense,
author = {Pan, Youmei and Wang, Peng and Hui, Xinyao and Li, Jinwen},
title = {Dense Points Aggregation for Efficient and Collaborative Earth-Imaging Task Planning},
year = {2023},
isbn = {9781450398336},
publisher = {Association for Computing Machinery},
address = {New York, NY, USA},
doi = {10.1145/3579654.3579705},
booktitle = {Proceedings of the 2022 5th ACAI},
articleno = {48},
numpages = {8},
location = {Sanya, China},
series = {ACAI '22}
}
\end{document}